\documentclass[sigconf]{acmart}

\usepackage[algoruled,linesnumbered]{algorithm2e}
\usepackage[font=small]{caption}
\usepackage{booktabs}
\usepackage[font=footnotesize]{subcaption}
\usepackage{multirow}
\usepackage{tabularx}
\usepackage{amsmath}
\usepackage{enumitem}
\usepackage{color}
\usepackage{xcolor,colortbl}
\usepackage{hhline}
\usepackage[export]{adjustbox}
\usepackage{tikz}
\usetikzlibrary{calc,arrows}
\usepackage{pgfplots}
\usepackage{pgfplotstable}
\usepackage{filecontents}
\usepgfplotslibrary{patchplots}
\usepgfplotslibrary{colormaps} 
\usetikzlibrary[pgfplots.colormaps,patterns] 
\usepackage{xstring}

\pgfplotsset{compat=newest,
    /pgfplots/ybar legend/.style={
    /pgfplots/legend image code/.code={%
       \draw[##1,/tikz/.cd,yshift=-0.25em]
        (0cm,0cm) rectangle (3pt,0.8em);},
   },
}

\hypersetup{pdfauthor={Author's name},%
pdftitle={Document Title},%
pagebackref=true,%
colorlinks=true,%
pdftex
}

\SetAlFnt{}
\SetAlgoSkip{}
\SetAlgoInsideSkip{}

\newcommand{\RN}[1]{%
  \textup{\uppercase\expandafter{\romannumeral#1}}%
}

\newcommand{\Rn}[1]{%
  \textit{\lowercase\expandafter{\romannumeral#1}}%
}

\newcommand{\citectl}[1]{%
   \edef\replace{:long,}%
   \edef\append{:long}%
   \StrSubstitute{#1}{,}{\replace}[\mystring]%
   \cite{\mystring\append}%
}%

\newcommand{\calprecent}[3]{\pgfplotstablegetelem{#2}{MM}\of{#1}%
\pgfmathsetmacro{\firstvalue}{\pgfplotsretval}%
\pgfplotstablegetelem{#2}{FP}\of{#1}%
\pgfmathsetmacro{\secondvalue}{\pgfplotsretval}%
\pgfmathsetmacro{#3}{(\firstvalue)/(\secondvalue)*100}}

\newcommand{\shiftprecent}[3]{\pgfplotstablegetelem{#2}{MM}\of{#1}%
\edef\firstvalue{\pgfplotsretval}%
\pgfplotstablegetelem{#2}{MMerror}\of{#1}%
\pgfmathparse{\firstvalue+\pgfplotsretval}%
\edef\firstsum{\pgfmathresult}%
\pgfplotstablegetelem{#2}{FP}\of{#1}%
\edef\secondvalue{\pgfplotsretval}%
\pgfplotstablegetelem{#2}{FPerror}\of{#1}%
\pgfmathparse{\secondvalue+\pgfplotsretval}%
\edef\secondsum{\pgfmathresult}%
\pgfmathparse{ifthenelse(\firstsum>\secondsum,\firstsum,\secondsum)}%
\edef#3{\pgfmathresult}}

\definecolor{lightblue}{RGB}{37,165,203}
\definecolor{darkblue}{RGB}{0,50,240}
\definecolor{darkyellow}{RGB}{253,177,26}
\definecolor{darkred}{HTML}{D7191C}
\definecolor{lightgreen}{HTML}{ABDDA4}
\definecolor{darkgreen}{RGB}{50,232,70}

\newcommand{\sysname}{Augur}

\begin{document}

\renewcommand\footnotetextcopyrightpermission[1]{} 

\copyrightyear{2017}
\acmYear{2017}
\setcopyright{usgovmixed}
\acmConference{MM'17}{}{October 23--27, 2017, Mountain View, CA, USA.}
\acmPrice{15.00}
\acmDOI{10.1145/3123266.3123389}
\acmISBN{978-1-4503-4906-2/17/10}

\fancyhead{}
\settopmatter{printfolios=false, printacmref=false}

\title[Modeling the Resource Requirements of CNNs on Mobile Devices]{Modeling the Resource Requirements of Convolutional Neural Networks on Mobile Devices}

\author{Zongqing Lu}
\orcid{0000-0003-3967-2704}
\affiliation{%
  \institution{Peking University}
}
\email{zongqing.lu@pku.edu.cn}

\author{Swati Rallapalli}
\affiliation{%
  \institution{IBM Research}
}
\email{srallapalli@us.ibm.com }

\author{Kevin Chan}
\affiliation{%
  \institution{Army Research Laboratory}
}
\email{kevin.s.chan.civ@mail.mil }

\author{Thomas La Porta}
\affiliation{%
  \institution{Pennsylvania State University}
}
\email{tlp@cse.psu.edu}

\renewcommand{\shortauthors}{Z. Lu et al.}

\begin{abstract}
Convolutional Neural Networks (CNNs) have revolutionized the research in computer vision, due to their ability to capture complex patterns, resulting in high inference accuracies. However, the increasingly complex nature of these neural networks means that they are particularly suited for server computers with powerful GPUs. We envision that deep learning applications will be eventually and widely deployed on mobile devices, \textit{e.g.}, smartphones, self-driving cars, and drones. Therefore, in this paper, we aim to understand the resource requirements (time, memory) of CNNs on mobile devices. First, by deploying several popular CNNs on mobile CPUs and GPUs, we measure and analyze the performance and resource usage for every layer of the CNNs. Our findings point out the potential ways of optimizing the performance on mobile devices. Second, we model the resource requirements of the different CNN computations. Finally, based on the measurement, profiling, and modeling, we build and evaluate our modeling tool, \emph{\sysname}, which takes a CNN configuration (descriptor) as the input and estimates the compute time and resource usage of the CNN, to give insights about whether and how efficiently a CNN can be run on a given mobile platform. In doing so \sysname~ tackles several challenges: (\Rn{1}) how to overcome profiling and measurement overhead; (\Rn{2}) how to capture the variance in different mobile platforms with different processors, memory, and cache sizes; and (\Rn{3}) how to account for the variance in the number, type and size of layers of the different CNN configurations. 
\end{abstract}


\maketitle

\section{Introduction}
Deep learning has become the norm of state-of-the-art learning systems, especially in computer version. 
Convolutional Neural Networks (CNNs) have demonstrated impressive performance on various computer vision tasks from classification and detection to segmentation and captioning. 
A CNN consists of different types of layers (\textit{e.g.}, convolutional, pooling, fully connected), where each layer performs certain transform on the input data and outputs the data to the next layer. 
Different CNNs for computer vision tasks have been designed, from a few layers to a thousand layers. 
But, the core of these networks naturally are the convolutional layers, which consist of a set of learnable kernels that are convolved across the length and width of the input image to produce output features. 
There are several frameworks that support the training (forward and backward pass) and inference (only forward pass) phases of CNNs, including Caffe \citectl{caffe}, TensorFlow \citectl{tensorflow}, Torch \citectl{torch}, Theano \citectl{theano}, \textit{etc}.
All of these frameworks are designed and optimized for both training and inference on computers with powerful GPUs.
 
However, we envision that deep learning applications will be eventually and widely deployed on mobile devices. 
It is also expected that for computer vision tasks mobile devices will only perform inference (forward pass), since training can be carried out offline by computers with powerful GPUs. 
In the rest of this paper, the terms ``inference", ``test" or ``forward pass'', mean the same.

Since both the frameworks, as well as the CNN models are designed for computers with powerful GPUs, they may not effectively and efficiently work on mobile devices due to several factors, \textit{e.g.}, constrained memory and limited computing capability.
CNNs for vision tasks are very complex -- for example, VGGNet \citectl{vgg} has 528M parameters and requires over 15G FLOPs (FLoating-point OPerations) to classify a single image. 
Due to the large amount of parameters and FLOPs, and the need to enable running these CNNs on resource-constrained mobile devices, 
several works focus on accelerating the computing of CNNs on mobile devices by compressing parameters \citectl{wu2016quantized,kim2016compression}, by cloud offload \citectl{han2016mcdnn}, and by distributing computation to heterogeneous processors on-board \citectl{lane2016deepx}. 
However, complementary to these techniques, our goal is to model the resource requirements of CNNs accurately. 
\textit{Motivation for this is that our system can serve guidelines to decide when performance optimizations, offloading, \textit{etc}. are required to successfully run analytics tasks on mobile devices. For instance, using the output of our models, one could decide to run all the convolutional layers on the mobile device while offloading the fully connected layers to the cloud so as to cut down on the memory requirement on the mobile device.} 
Although accurately modeling the resource requirements of CNNs is very hard, we make progress towards achieving it.

This paper overviews the workflow of CNNs, shares the experiences of deploying CNNs on mobile devices, gives the performance measurements and analysis, and models the resource requirements of the inference phase of the CNNs on mobile devices.
In doing so we face significant challenges. (\Rn{1}) \emph{Profiling overhead}: to measure timing of GPU computations,
we need to add a synchronization call that waits for all the results to come back before recording the time. 
As pointed out by \citectl{cuda-guide}, this causes an overhead, as some cores may be idling while waiting for the rest of the cores to complete the computation. 
We address this challenge by amortizing this measurement cost by executing the computing task a large number of times and averaging the running time. This ensures that
the overhead per iteration is negligible.
(\Rn{2}) \emph{Different types of layers}: CNNs are composed of various types of layers, so to model the resource requirements of all the different types is challenging. 
On the other hand, since main computation of all these layers boils down to matrix multiplication, we are able to model the different layers by abstracting out the details and focusing on the core of the computation.
(\Rn{3}) \textit{How matrix multiplication scales}: as the core of the computation of CNNs, it is important to understand how the computation scales with the sizes of matrices in terms of the resource requirements. 
Due to the large number of combinations of matrix sizes, this can be very challenging. 
However, by extracting the matrix multiplication sizes of popular CNNs, we observe that all of them result into a small set of matrix sizes and thus we are able to accurately model them for different mobile platforms.

\textbf{Contributions:} (\Rn{1}) We deploy the popular CNN models including AlexNet \citectl{alexnet}, VGGNet \citectl{vgg}, GoogleNet \citectl{google}, and ResNet \citectl{resnet} using the Caffe framework \citectl{jia2014caffe} on mobile platforms (\textit{i.e.}, NVIDIA TK1 and TX1), where the inference phase is run on both CPUs and GPUs (\S\ref{sec:deployment}).
(\Rn{2}) We measure and analyze the performance and resource usage of the inference phase of these CNN models on a layerwise granularity. 
Our findings point out the potential ways of optimizing the computing of CNNs on mobile devices (\S\ref{sec:measure}). (\Rn{3}) We profile and model the resource requirements of CNNs. 
We also build a modeling tool, \emph{\sysname}, which takes a CNN model descriptor as the input and estimates the resource requirements of the CNN so as to give insights on how well the CNN can be run on a mobile platform without having to implement and deploy it (\S\ref{sec:profile}). 

\section{Background}

\subsection{Overview of CNNs}       
\label{ssec:overview}
Our goal is to model the resource requirements of the forward pass of a CNN. 
The CNN architecture is typically composed of convolutional, normalization, and subsampling layers optionally followed by fully connected layers.
We overview these layers below, as it lays the foundations for modeling the resource requirements. 

\noindent \textbf{Convolutional Layer:} The convolutional (CONV) layers form the core of CNNs. 
The parameters of this layer are a set of kernels (weights) and biases learned during the training phase.
During the forward pass, kernels are convolved across the width, height, and depth of the input, computing the dot product between the kernel and the input and producing the output volume. 
Since the main operation is dot product between the kernels and local regions of the input, the forward pass of a CONV layer can be formulated as a matrix multiplication.
For the input volume, each local region (a block of pixels) is stretched into a column of a matrix, and the number of columns is the total number of local regions. 
The kernel is stretched into a column of another matrix, and the number of columns is the number of kernels.
Finally, the product of the matrix multiplication is reshaped to the output volume with a depth equal to the number of kernels. 
For example, the input of AlexNet $[227 \times 227 \times 3]$ (width $\times$ height $\times$ depth) is convolved with 
96 kernels at size $[11 \times 11 \times 3]$ and with a stride 4, and hence there are 55 locations along both width and height. 
So, the matrix for the input is $[3025 \times 363]$, the matrix of 
the kernels is $[363 \times 96]$, and the produced matrix is $[3025 \times 96]$ and finally reshaped 
to $[55 \times 55 \times 96]$. 

The CONV layer is commonly implemented using the matrix multiplication function of Basic Linear Algebra Subprograms (BLAS) 
on CPUs and cuBLAS \citectl{cublas} on CUDA GPUs for acceleration. However, as many values in the input volume are 
replicated multiple times in the matrix stretched from the input volume, it uses more memory than the input volume itself.

\noindent \textbf{Pooling Layer:} The pooling (POOL) layer commonly sits between CONV layers and performs downsampling 
to reduce the spatial size (width and height).
The pooling is performed on local regions with the kernel size defined by a CNN model. The most common pooling operation in the state-of-the-art CNN models is max pooling. 
The pooling layer independently operates on the input volume without parameters, and hence its implementation is simple.

\noindent \textbf{Normalization Layer:} Two types of normalization layers are commonly used in CNNs: local response normalization (LRN) and batch normalization (BatchNorm). 
However, LRN's role has been outperformed by other techniques, such as BatchNorm, and thus here we only detail BatchNorm.

BatchNorm is introduced to reduce the internal covariant shift during training \citectl{batchnorm}.
During test phase, BatchNorm normalizes the input volume on each dimension (weight $\times$ height), \textit{e.g.}, for the $i$-th dimension, as follows,
\begin{equation}
\label{eq:bn}
\nonumber\widehat{x}^{(i)} =  \frac{x^{(i)} - \mathrm{E}[x^{(i)}]}{\sqrt{\mathrm{Var}[x^{(i)}]}},
\end{equation}
\noindent where $\mathrm{E}[x^{(i)}]$ and $\mathrm{Var}[x^{(i)}]$ are learned during the training phase for dimension $i$. 

\noindent \textbf{Fully Connected Layer:} 
Each neuron in a fully connected (FC) layer is connected to all activations in the previous layer. 
Due to the full connectivity, there are a huge number of parameters, which places heavy burden on memory usage and computation.
Recently, FC layers have fallen out of favor, \textit{e.g.}, the latest CNNs, \textit{i.e.}, GoogleNet and ResNet, only have one fully connected layer as the classifier. 
This dramatically reduces the number of parameters, \textit{e.g.}, 26MB parameters in GoogleNet while 233MB in AlexNet. Moreover, it was found that FC layers of VGGNet can be removed with no performance reduction. 
Therefore, it is anticipated that CNNs will eliminate the use of FC layers.
The forward pass of FC layers is also implemented as a matrix multiplication. 

Besides these four layers, rectified linear unit (ReLU) layer that applies an elementwise function, \textit{e.g.}, $\max(0,x)$, on the input volume, is also commonly used in CNNs.
However, ReLU is simple, has no parameters, and does not change the size of input volume. 
Thus we skip the detail of ReLU layer.

\begin{table*}
\centering
\begin{minipage}{0.33\textwidth}
\setlength{\abovecaptionskip}{2pt}
\renewcommand{\arraystretch}{1.2}
\caption{CNN models}
\label{tab:models}
\centering
\begin{scriptsize}
\begin{tabular}{ccccc}
\hline
Layer & AlexNet & VGGNet & GoogLeNet & ResNet \\ \hline
CONV & 5 & 13 & 57 & 53 \\ 
POOL & 3 & 5 & 14 & 2 \\ 
NORM & 2 &  & 2 & 53 \\ 
ReLU & 7 & 15 & 57 & 49 \\ 
FC & 3 & 3 & 1 & 1 \\ 
Concat & \multirow{3}{*}{} & \multirow{3}{*}{} & 9 &  \\
Scale &  &  &  & 53 \\
Eltwise &  &  &  & 16 \\ \hline
Total & 20 & 36 & 140 & 227 \\ \hline
\end{tabular}
\end{scriptsize}
\end{minipage}
\hspace{0.4cm}
\begin{minipage}{0.58\textwidth}
\setlength{\abovecaptionskip}{2pt}
\renewcommand{\arraystretch}{1.2}
\centering
\caption{Timing benchmarks on AlexNet}
\label{tab:alexnet}
\begin{scriptsize}
\begin{tabular}{ccccccccc}
\hline
\multicolumn{2}{c}{\multirow{2}{*}{Platform}} & \multicolumn{5}{c}{Layerwise Pass (ms)} & 
\multirow{2}{*}{Total (ms)} & \multirow{2}{*}{Forward Pass (ms)} \\
\multicolumn{2}{c}{} & CONV & POOL & LRN & ReLU & FC & &\\  \hline
\multirow{4}{*}{TK1} & \multirow{2}{*}{CPU} & 318.7$\pm$0.2 & 6.1$\pm$0.1 & 103.8$\pm$0.0 & 4.6$\pm$0.0 & 186.3$\pm$0.1 & \multirow{2}{*}{619.8$\pm$0.2} & \multirow{2}{*}{619.5$\pm$0.2} \\
 &  & {\bf 51.42}\% & 0.99\% & 16.74\% & 0.75\% & 30.05\% &  &  \\ 
 & \multirow{2}{*}{GPU} & 24.6$\pm$3.5 & 2.3$\pm$0.6 & 2.4$\pm$0.5 & 5.2$\pm$1.2 & 35.1$\pm$5.9 & \multirow{2}{*}{73.3$\pm$10.7} & \multirow{2}{*}{54.7$\pm$2.4} \\
 &  & 33.53\% & 3.15\% & 3.22\% & 7.11\% & {\bf 47.95}\% &  &  \\ 
\multirow{4}{*}{TX1} & \multirow{2}{*}{CPU} & 66.9$\pm$5.3 & 7.6$\pm$0.0 & 172.4$\pm$0.3 & 2.4$\pm$0.0 & 644.7$\pm$5.3 & \multirow{2}{*}{894.3$\pm$4.8} & \multirow{2}{*}{892.7$\pm$2.3} \\
 &  & 7.48\% & 0.85\% & 19.28\% & 0.27\% & {\bf 72.09}\% &  &  \\ 
 & \multirow{2}{*}{GPU} & 24.2$\pm$8.3 & 1.3$\pm$2.6 & 2.7$\pm$3.0 & 5.9$\pm$5.9 & 15.2$\pm$4.7 & \multirow{2}{*}{52.8$\pm$15.7} & \multirow{2}{*}{29.3$\pm$6.5} \\
 &  & {\bf 45.79\%} & 2.51\% & 5.12\% & 11.23\% & 28.76\% &  &  \\ \hline
\multicolumn{2}{c}{\multirow{2}{*}{FLOPs}} & 666M & 1M & 2M & 0.7M & 59M & \multicolumn{2}{c}{\multirow{2}{*}{729M}} \\
\multicolumn{2}{c}{} & {\bf 91.36\%} & 0.14\% & 0.27\% & 0.10\% & 8.09\% & \multicolumn{2}{c}{} \\ \hline
\end{tabular}
\end{scriptsize}
\end{minipage}
%
\end{table*}

\begin{table*}
\centering
\begin{minipage}{0.56\textwidth}
\setlength{\abovecaptionskip}{2pt}
\centering
\renewcommand{\arraystretch}{1.2}
\caption{Timing benchmarks on VGGNet}
\label{tab:vgg}
\begin{scriptsize}
\begin{tabular}{cccccccc}
\hline
\multicolumn{2}{c}{\multirow{2}{*}{Platform}} & \multicolumn{4}{c}{Layerwise Pass (ms)} & \multirow{2}{*}{Total (ms)} & \multirow{2}{*}{Forward Pass (ms)} \\ 
\multicolumn{2}{c}{} & CONV & POOL & ReLU & FC &  & \\ \hline
\multirow{4}{*}{TK1} & \multirow{2}{*}{CPU} & 7160.5$\pm$0.7 & 60.1$\pm$0.1 & 95.6$\pm$0.1 & 381.6$\pm$0.2 & \multirow{2}{*}{7697.9$\pm$0.6} & \multirow{2}{*}{7697.8$\pm$0.5} \\
 &  & {\bf 93.02\%} & 0.78\% & 1.24\% & 4.96\% &  &  \\ 
 & \multirow{2}{*}{GPU} & 263.1$\pm$19.3 & 7.2$\pm$0.5 & 17.5$\pm$1.2 & 57.6$\pm$0.5 & \multirow{2}{*}{347.6$\pm$20.1} & \multirow{2}{*}{326.7$\pm$2.1} \\
 &  & {\bf 75.68\%} & 2.06\% & 5.03\% & 16.58\% &  &  \\
\multirow{4}{*}{TX1} & \multirow{2}{*}{CPU} & 1952.9$\pm$12.2 & 71.3$\pm$1.5 & 52.5$\pm$1.9 & 747.7$\pm$24.9 & \multirow{2}{*}{2824.6$\pm$23.2} & \multirow{2}{*}{2809.1$\pm$10.6} \\
 &  & {\bf 69.14\%} & 2.52\% & 1.86\% & 26.47\% &  &  \\ 
 & \multirow{2}{*}{GPU} & 136.3$\pm$5.4 & 3.4$\pm$1.6 & 9.9$\pm$4.9 & 32.8$\pm$1.3 & \multirow{2}{*}{184.2$\pm$7.4} & \multirow{2}{*}{175.3$\pm$2.0} \\
 &  & {\bf 73.98\%} & 1.84\% & 5.35\% & 17.82\% &  &  \\ \hline
\multicolumn{2}{c}{\multirow{2}{*}{FLOPs}} & 15360M & 6M & 14M & 124M & \multicolumn{2}{c}{\multirow{2}{*}{15503M}} \\
\multicolumn{2}{c}{} & {\bf 99.08\%} & 0.04\% & 0.09\% & 0.79\% & \multicolumn{2}{c}{} \\ \hline
\end{tabular}
\end{scriptsize}
\end{minipage}
\hspace{0.3cm}
\begin{minipage}{0.37\textwidth}
\setlength{\abovecaptionskip}{2pt}
\renewcommand{\arraystretch}{1.2}
\centering
\renewcommand{\arraystretch}{1.2}
\caption{Memory of CNN models on platforms (MB)}
\label{tab:memory}
\begin{scriptsize}
\begin{tabular}{cccccc}
\hline
\multicolumn{2}{c}{Type/Platform}     & AlexNet & VGGNet & GoogleNet & ResNet \\ \hline
\multicolumn{2}{c}{Weights \& Biases}  & 233  & 528  & 26   & 97       \\ 
\multicolumn{2}{c}{Data} & 8   & 110   & 53  & 221      \\ 
\multicolumn{2}{c}{Workspace}  & 11  & 168  & 46  & 79       \\ \hline 
\multirow{2}{*}{TK1}  & CPU   & 324  & 972   & 161  & 409   \\ 
  & GPU     & 560   & 1508  & 196       & 533      \\ 
\multirow{2}{*}{TX1}  & CPU   & 362   & 1013  & 200  & 453   \\ 
  & GPU     & 589   & 1537  & 226  & 562   \\ \hline
\end{tabular}
\end{scriptsize}
\end{minipage}
%
\end{table*}

\subsection{Related Work}
\label{ssec:related}
Although CNNs have been applied to various computer vision applications on different computing platforms, only a few works consider running CNNs on mobile devices, which we envision to be a significant future area for the deployment of deep learning applications.

Among these works, many focus on accelerating the computing of CNNs, \textit{e.g.}, by compressing parameters \citectl{wu2016quantized,kim2016compression}, by cloud offload \citectl{han2016mcdnn}, and by distributing computation to heterogeneous processors on-board \citectl{lane2016deepx}. 
Some consider reducing the memory usage to better fit mobile devices while maintaining high inference accuracy, \textit{e.g.}, \citectl{gong2014compressing,iandola2016squeezenet}. The resource bottlenecks of running CNNs on mobile devices are preliminarily investigated in \citectl{lane2015early}.
Different CNNs are benchmarked in \citectl{canziani2016analysis}, but it does not consider how to model the resource requirements of CNNs.

While CNNs grow from a few layers to a thousand layers, the computational capability of mobile devices continues to improve.
As a result, different mobile devices perform differently on different CNNs, and hence custom optimization and offloading may or may not be needed. 
It depends on whether and how efficiently a CNN can be run on a given mobile platform.
This question motivates our work. 

\section{Measurement Set-Up}
\label{sec:deployment}
To understand the resource requirement of the forward pass of CNNs, we deployed several CNN models on two mobile platforms using the popular deep learning framework -- Caffe.

\noindent \textbf{Platforms:} Although some frameworks (\textit{e.g.}, Caffe, Torch) can run on Android and iOS, they do not support GPU acceleration on off-the-shelf mobile devices, such as smartphones or tablets. 
To understand the performance of CNNs on both mobile CPUs and GPUs, in this paper, we focus on two developer kits for low power edge devices -- NVIDIA TK1 and TX1.

TK1 is equipped with a 2.3GHz quad-core ARM Cortex-15A 32bit CPU, 192 CUDA cores Kepler GPU, and 2GB DDR3L RAM.
TX1 is more powerful and has a 1.9GHz quad-core ARM Cortex-A57 64bit CPU, 256 CUDA cores Maxwell GPU, and 4GB LPDDR4 RAM. 
The system-on-chip (including CPU and GPU) of TK1 and TX1 also appears in many off-the-shelf mobile devices, such as Google Nexus 9 and Pixel C. 
However, none of these devices are enabled to support CUDA, on which deep learning frameworks are built for GPU acceleration.
Thus, for ease of experimentation we choose NVIDIA TK1 and TX1, the results of which should indicate the performance of CNNs on mobile devices.         

\noindent \textbf{Framework:} There are several frameworks for deep neural networks.
As mentioned before, most of the frameworks use BLAS on CPU and cuBLAS on GPUs for the CNN computations and thus show similar performance.
In this paper, we use the popular Caffe framework, where the choice of BLAS is OpenBLAS \citectl{openblas}.

\noindent \textbf{CNN Models:} For the measurement, we consider the most popular CNN models including AlexNet, VGGNet (VGG-16), GoogleNet, and ResNet (ResNet-50). 
Although the architectures of these models are quite different, from several layers to more than one hundred layers and from regular stacked layers to branched and stacked layers, they are mainly built on the basic layers of CNNs. 
Table~\ref{tab:models} shows how many these layers each model contains.

\section{Initial Measurement Study}
\label{sec:measure}
In this section, we investigate the resource requirements and bottlenecks of running several 
well known CNN models on mobile platforms.

\begin{table*}
\centering
\setlength{\abovecaptionskip}{2pt}
\renewcommand{\arraystretch}{1.2}
\caption{Timing benchmarks on GoogleNet}
\label{tab:googlenet}
\begin{scriptsize}
\begin{tabular}{cccccccccc}
\hline
\multicolumn{2}{c}{\multirow{2}{*}{Platform}} & \multicolumn{6}{c}{Layerwise Pass (ms)} & \multirow{2}{*}{Total (ms)} & \multirow{2}{*}{Forward Pass (ms)} \\ 
\multicolumn{2}{c}{} & CONV & POOL & LRN & ReLU & Concat & FC &  & \\ \hline
\multirow{4}{*}{TK1} & \multirow{2}{*}{CPU} & 755.3$\pm$0.2 & 68.8$\pm$0.1 & 214.3$\pm$0.2 & 22.8$\pm0.0$ & 2.0$\pm$0.0 & 2.7$\pm$0.0 & \multirow{2}{*}{1066.2$\pm$0.3} & \multirow{2}{*}{1065.6$\pm$0.2} \\
 &  & {\bf 70.84\%} & 6.45\% & 20.10\% & 2.14\% & 0.19\% & 0.26\% &  &  \\ 
 & \multirow{2}{*}{GPU} & 186.9$\pm$45.0 & 20.6$\pm$4.9 & 6.5$\pm$1.5 & 35.3$\pm$9.9 & 13.0$\pm$4.4 & 2.4$\pm$0.8 & \multirow{2}{*}{269.3 $\pm$65.6} & \multirow{2}{*}{167.0$\pm$44.3} \\
 &  & {\bf 69.40\%} & 7.65\% & 2.40\% & 13.10\% & 4.81\% & 0.90\% &  &  \\ 
\multirow{4}{*}{TX1} & \multirow{2}{*}{CPU} & 174.4$\pm$3.6 & 89.9$\pm$0.2 & 349.4$\pm$0.6 & 9.5$\pm$0.1 & 1.7$\pm$0.1 & 5.7$\pm$0.0 & \multirow{2}{*}{630.9$\pm$3.5} & \multirow{2}{*}{637.9$\pm$14.7} \\
 &  & 27.64\% & 14.24\% & {\bf 55.38\%} & 1.50\% & 0.36\% & 0.90\% &  &  \\ 
 & \multirow{2}{*}{GPU} & 165.9$\pm$48.8 & 18.5$\pm$11.2 & 3.3$\pm$2.3 & 49.5$\pm$31.2 & 15.4$\pm$9.8 & 1.2$\pm$1.1 & \multirow{2}{*}{258.1$\pm$89.8} & \multirow{2}{*}{143.9$\pm$59.2} \\
 &  & {\bf 64.28\%} & 7.16\% & 1.28\% & 19.16\% & 5.96\% & 0.46\% &  &  \\ \hline
\multicolumn{2}{c}{\multirow{2}{*}{FLOPs}} & 1585M & 13M & 3M & 3M &  & 1M & \multicolumn{2}{c}{\multirow{2}{*}{1606M}} \\
\multicolumn{2}{c}{} & {\bf 98.80\%} & 0.80\% & 0.20\% & 0.20\% &  & 0.06\% & \multicolumn{2}{c}{} \\ \hline
\end{tabular}
\end{scriptsize}
\end{table*}

\begin{table*}
\setlength{\abovecaptionskip}{2pt}
\centering
\renewcommand{\arraystretch}{1.2}
\caption{Timing benchmarks on ResNet}
\label{tab:resnet}
\begin{scriptsize}
\begin{tabular}{ccccccccccc}
\hline
\multicolumn{2}{c}{\multirow{2}{*}{Platform}} & \multicolumn{7}{c}{Layerwise Pass (ms)} & \multirow{2}{*}{Total (ms)} & \multirow{2}{*}{Forward Pass (ms)} \\ 
\multicolumn{2}{c}{} & CONV & POOL & BatchNorm & ReLU & Scale & Eltwise & FC &  & \\ \hline
\multirow{4}{*}{TK1} & \multirow{2}{*}{CPU} & 1830.4$\pm$0.4 & 8.8$\pm$0.0 & 97.1$\pm0.1$ & 64.0$\pm$0.1 & 42.0$\pm$0.1 & 24.8$\pm$0.1 & 5.4$\pm$0.0 & \multirow{2}{*}{2072.7$\pm$0.4} & \multirow{2}{*}{2072.2$\pm$0.3} \\
 &  & {\bf 88.31\%} & 0.42\% & 4.68\% & 3.09\% & 2.03\% & 1.20\% & 0.26\% &  &  \\ 
 & \multirow{2}{*}{GPU} & 245.8$\pm$16.3 & 5.5$\pm$0.6 & 249.5$\pm$11.6 & 38.7$\pm$2.0 & 76.0$\pm$3.3 & 47.0$\pm$2.7 & 3.9$\pm$0.1 & \multirow{2}{*}{673.0$\pm$33.4} & \multirow{2}{*}{149.4$\pm$4.9} \\
 &  & 36.53\% & 0.81\% & {\bf 37.08\%} & 5.75\% & 11.29\% & 6.98\% & 0.58\% &  &  \\ 
\multirow{4}{*}{TX1} & \multirow{2}{*}{CPU} & 362.3$\pm$5.4 & 13.7$\pm$0.2 & 83.5$\pm$0.3 & 33.2$\pm$0.1 & 31.9$\pm$3.6 & 20.4$\pm$4.2 & 22.2$\pm$0.1 & \multirow{2}{*}{567.6$\pm$7.6} & \multirow{2}{*}{566.8$\pm$9.7} \\
 &  & {\bf 63.83\%} & 2.41\% & 14.7\% & 5.86\% & 5.62\% & 3.59\% & 3.92\% &  &  \\ 
 & \multirow{2}{*}{GPU} & 279.4$\pm$42.6 & 3.0$\pm$2.7 & 198.1$\pm$36.8 & 63.6$\pm$31.3 & 79.8$\pm$24.2 & 34.8$\pm$12.9 & 1.8$\pm$2.4 & \multirow{2}{*}{664.7$\pm$116.5} & \multirow{2}{*}{104.4$\pm$14.0} \\
 &  & {\bf 42.03\%} & 0.45\% & 29.80\% & 9.57\% & 12.01\% & 5.24\% & 0.27\% &  &  \\ \hline
\multicolumn{2}{c}{\multirow{2}{*}{FLOPs}} & 3866M & 2M & 32M & 9M & 11M & 6M & 2M & \multicolumn{2}{c}{\multirow{2}{*}{3922M}} \\
\multicolumn{2}{c}{} & {\bf 98.59\%} & 0.05\% & 0.81\% & 0.23\% & 0.27\% & 0.14\% & 0.05\% & \multicolumn{2}{c}{} \\ \hline
\end{tabular}
\end{scriptsize}
\end{table*}

\subsection{Timing}
\label{sec:timing}
First, we measure the timing of each model on different platforms using CPU and GPU in terms of (\Rn{1}) complete forward pass: \emph{i.e.}, timing is measured for the entire forward pass and (\Rn{2}) as summation of individual layer times.
We also calculate the number of FLOPs for each model and each type of layer.  

\noindent \textbf{AlexNet} has the least number of layers among these models and indeed requires the least amount of computation in terms of FLOPs, \textit{i.e.}, 729M. 
As shown in Table~\ref{tab:alexnet}, on the CPU of both TK1 and TX1, the summation of layerwise timing perfectly matches with that of a full forward pass, which are about 600ms (on TK1) and 900ms (on TX1).
\textit{Surprisingly, although TX1 has a more powerful CPU, the forward pass on TX1 is slower than TK1}. 
The CONV layers on TX1 run much faster than on TK1 (more than 4x), but the FC layers are much slower (more than 3x).
Since the basic computation of both CONV and FC is matrix multiplication, the results seem contradictory at first. However, we investigate and explain the reasons for the behavior below.

First, even though the clock is slower on TX1 compared to TK1, \textit{i.e.}, 1.9 GHz vs. 2.3 GHz, TX1 runs more instructions per clock cycle compared to TK1 (3 vs. 2) and hence the performance of TX1 CPU is expected to be better than TK1 CPU as we see for the CONV layers.
Second, FC layers have many more parameters than the CONV layers.
Therefore, FC layers are bottlenecked by the memory whereas CONV layers are compute bound.
Third, the L1 data cache size is 32 KB on both and L2 cache is larger on TK1 compared to TX1.
Even if cache size is same on both -- because the address is longer on TX1 (64 bit vs. 32 bit), more memory is used up for the addressing and we have lesser memory available to save the data itself on the cache. 
This means that we need to fetch data from RAM to the cache more often while executing the FC layers on TX1 due to the large number of parameters which causes the slow down. 

GPUs can significantly accelerate the computation of a CNN and thus improve the performance over CPUs.
More advanced TX1 GPU outperforms TK1 GPU as expected. 
However, we face one challenge: the summation of layerwise timing does not match the timing of the full forward pass on GPUs. 
The reason for the mismatch is that CUDA supports asynchronous programming. 
Before time measurement, an API (cudaDeviceSynchronize) has to be called to make sure that all cores have finished their tasks. This explicit synchronization is the overhead of measuring time on the GPUs. 
Therefore, the sum of layerwise timing on GPUs is longer than a full forward pass. 

\noindent \textbf{VGGNet} has 2x CONV layers compared to AlexNet (Table~\ref{tab:models}). 
However, the number of operations is 20x that of AlexNet because VGGNet uses much larger feature maps. 
While other results follow similar pattern as AlexNet, the throughput of both CPU and GPU on VGGNet is higher than on AlexNet.
For example, the throughput of TK1 CPU on AlexNet is 1 GFLOPS (GFLOPs per Second) and of VGGNet is 2 GFLOPS. This is mainly because both CPU and GPU have better throughput on matrix multiplication with larger size. 

\noindent \textbf{GoogleNet} has more than 50 CONV layers, many more than AlexNet. 
However, the CONV layers have only two times more FLOPs than that of AlexNet.
The main reason is that the size of the kernels and feature maps is small, which dramatically reduces the number of operations.
Similar to AlexNet, GoogleNet also employs LRN that significantly affects the performance on CPU for both TK1 and TX1. For example, it takes more than 55\% of total time on TX1 CPU. 
GoogleNet has a layer, named Concat, that does not involve any computation, but concatenates the outputs from previous layers, thus involving memory operations only. 

The difference between layerwise timing and full forward pass on GoogleNet is much larger than AlexNet and VGGNet as shown in Table~\ref{tab:googlenet}.
GoogleNet has many more layers than AlexNet and VGGNet and thus much more measuring overhead on GPUs.
The measuring overhead may be larger than the compute time when the computation of a layer does not cost much time, \textit{e.g.}, ReLU layers. 
Due to this measurement artifact, in Table~\ref{tab:googlenet}, ReLU layers cost more time on GPUs than CPUs. 
This is a motivation for us to devise measurement techniques that can overcome these measurement overheads as we see later in \S\ref{sec:profile}. 

\noindent \textbf{ResNet} has more than two hundred layers. 
ResNet includes BatchNorm, Scale, and Eltwise that are not commonly used by other models. 
These layers are not expensive in terms of FLOPs as shown in Table~\ref{tab:resnet}. 
We observe that the computation of Scale and Eltwise costs more on GPUs than CPUs, which is again due to the measurement overhead on GPUs as discussed above. 
Interestingly, although ResNet has more FLOPs (2x) than GoogleNet, a full forward pass is faster than GoogleNet on TX1. This is because LRN of GoogleNet is very expensive: (55\% of total time) on TX1 CPU. 
Moreover, GoogleNet has more CONV layers and the underlying matrix multiplication is smaller than that of ResNet. 
As GPU throughput is higher on matrix multiplication with larger size, ResNet is faster than GoogleNet on TX1 GPU.

\subsection{Memory}
\label{ssec:memory}
The memory requirement to run a CNN comes from three major sources: (\Rn{1}) the memory that holds the parameters of the CNN; (\Rn{2}) the memory that stores intermediate data of the CNN; and (\Rn{3}) the workspace for computation.
A majority of the CNN parameters come from CONV and FC layers (\textit{i.e.}, weights and biases).
Intermediate data is the output of each layer (\textit{i.e.}, the input of next layer), \textit{e.g.}, feature maps.
Some types of layers require additional space to perform computation, \textit{e.g.}, on CONV layers, the memory is needed to hold the matrix stretched from the input data for matrix multiplication.
The workspace memory is mostly consumed by the matrix multiplication of CONV layers.
The NVIDIA CUDA Deep Neural Network library (cuDNN) \citectl{cudnn} can reduce the workspace by sacrificing the speed of computing on GPUs.
However, as the workspace is not the most significant part, cuDNN cannot reduce the memory usage of CNNs significantly. 

Table~\ref{tab:memory} shows the memory requirement of weights and biases of CONV and FC layers, intermediate data, and workspace of CONV layers for each CNN -- by parsing the model descriptor (\textit{e.g.}, a prototxt file in Caffe).
Table~\ref{tab:memory} also gives the measured memory usage of Caffe, running each CNN on these platforms. 
One can see that deeper CNNs (from AlexNet to ResNet) may not require more memory, especially for GoogleNet, which requires the least memory among them. 
Memory usage on TX1 is more than TK1, because TK1 is running a 32-bit OS while TX1 is running a 64-bit OS, which incurs more memory usage for the framework itself.

To speed up the computation of CNNs, all memory should be allocated beforehand and not released during the computation. Although existing frameworks (\textit{e.g.}, Caffe\footnote{Caffe allocates the memory for intermediate data on demand (lazily) during the first run, and thus it takes longer time than later runs.}) follow this rule, they are designed for training and testing (scoring) on workstations with powerful GPUs, and thus not quite suitable for mobile devices in terms of memory management. 

\noindent\textbf{Unified Memory Architecture:} Unlike workstations\footnote{Although GPUs on workstations can also directly access host memory over PCIe, \textit{e.g.}, CUDA kernels, reading data over PCIe is limited by PCIe bandwidth (up to 32GB/s) which is much slower than reading data from GPU memory (limit 200GB/s).} where GPUs have their dedicated memory, mobile platforms usually have a unified memory architecture, where GPU shares system memory with CPU.
On workstations, in the current implementation of Caffe, data is transferred to and from the memory of GPU for access, which is efficient on workstations.
However, on united memory architecture, \textit{e.g.}, TK1 and TX1, memory transfer from CPU to GPU simply generates a redundant data copy on system memory.
As shown in Table~\ref{tab:memory}, on both TK1 and TX1, the memory usage on GPU is always more than CPU, and the additional memory is actually used to hold a redundant copy of the parameters of each CNN (mostly weights and biases).
For example, running AlexNet on TK1 GPU takes 560MB memory, which is 236MB more than TK1 CPU, while weights and biases of AlexNet are 233MB in total. This also stands for other CNNs.

Mobile GPUs can directly access data by mapping host memory without degrading performance and incurring memory transfer overhead (\textit{i.e.}, \textit{zero-copy} memory). Existing frameworks, including Caffe, Torch, and Theano, do not take into consideration the unified memory architecture for mobile platforms. On the contrary, the unified memory architecture can be exploited to design a tailored computing framework for mobile devices. (\Rn{1}) We can eliminate memory transfers between CPU and GPU. (\Rn{2}) We can compute a CNN in the most efficient way; \textit{i.e.}, each layer can be executed on the most efficient unit, switching back and forth between GPU and CPU, without incurring additional memory transfer overhead.

\subsection{Analysis}
\textbf{FLOPs.} As the throughput of both CPU and GPU is higher on the CNN with more FLOPs and a significant amount of memory operations are involved in a CNN computation, FLOPs cannot accurately reflect the compute time of a CNN. 
For example, ResNet is faster than GoogleNet on GPUs, though it involves more FLOPs. 
Therefore, estimating the compute time of CNNs directly from  
their FLOPs is not feasible.

\noindent \textbf{CONV and FC Layer.} The computation of CONV and FC layers in most models 
accounts for a majority of FLOPs.
Therefore, can one measure these layers instead of the entire network?
However, this approach encounters other difficulties, \textit{i.e.}, layerwise measuring overhead on GPUs, and we have no way to know the exact overhead for each layer, which is hidden by GPUs.

\noindent \textbf{Matrix Multiplication.} The core of CONV and FC layers are matrix multiplications. 
Therefore, rather than going into the details of each of the individual layers, if we are able to extract
the matrix multiplication part of the layer, we will be able to accurately capture the resource requirements
of these layers.

\input{cpu_mm_fig.tex}

\section{\sysname}
\label{sec:profile}
We aim to build a modeling tool that can estimate the resource requirements of any given CNN descriptor on specific mobile platforms without implementation and deployment. 
This way, we can take the costs into consideration during the design of a CNN. This is critical when designing CNNs for resource-constrained mobile devices. 

\subsection{Profiling}
\label{sec:profiling}
The basic idea is simple. We first find the matrix multiplications that form the core of the CNN computation.
Then we measure their performance based on the BLAS and cuBLAS libraries, which are commonly 
used for matrix multiplications on CPUs and GPUs respectively.

\noindent\textbf{Extract matrix sizes:} To find all matrix multiplications and their sizes, we need to parse the descriptor of a CNN.
The dimension of input (\textit{e.g.}, images and feature maps) and network parameters (\textit{e.g.}, convolution kernels) determines two matrix sizes (that are to be multiplied) 
at a CONV or FC layer. 
As the dimension of feature maps can be changed by some other layers, \textit{e.g.}, POOL layers, we need to trace the dimension of feature maps layer by layer. 
However, this can be easily done by parsing the parameter settings at each layer, such as zero-padding ($P$), stride ($S$), the number of output feature maps ($N$). 
For instance, in case of a CONV layer, let $I$ denote the spatial dimension of the input feature map, $O$ denote the spatial dimension of the output feature map, $K$ denote the 3D volume of the convolution kernels. Then, we have:
\begin{align*}
O_w & = \lfloor (I_w-K_w+2P)/S \rfloor + 1 \\
O_h & = \lfloor (I_h-K_h+2P)/S \rfloor + 1. 
\end{align*}
Then, the matrix multiplication at the CONV layer is $[(O_w \cdot O_h) \times (K_w \cdot K_h \cdot K_d)][(K_w \cdot K_h \cdot K_d) \times N]$.
 
\noindent\textbf{Mitigate measurement overhead:} Layerwise timing measurement incurs heavy overhead on GPUs and causes a large deviation from a full forward pass. 
Moreover, the overhead is not fixed and varies over each measurement. 
As illustrated in Table~\ref{tab:googlenet} and \ref{tab:resnet}, the measurement overhead (the difference between the sum of layerwise measurements and full forward pass) of GoogleNet (131 measurements) on TX1 GPU is 128 ms, while the overhead of ResNet (227 measurements) is 595 ms. 
Therefore, we need a way to mitigate the overhead for accurate timing of matrix multiplications.

Timing measurements on GPUs can only been recorded after all cores finish their tasks. 
In a full forward pass, timing is only recorded at the last layer. 
Therefore, a core may be assigned with the computation of following layers and thus it can continuously perform the computation without synchronization. 
For example, after finishing the multiply-add operations for the matrix multiplication at a CONV layer, a core can continue to calculate the \textit{max} function of next ReLU layer on the output of multiply-add operations.
If layerwise timing is recorded, all cores have to wait until all multiply-add operations of the CONV layer have been completed.

The idea of mitigating the measurement overhead is simple.
To benchmark a matrix multiplication, we keep GPUs iteratively running the matrix multiplication in a way that GPU cores can continuously perform multiply-add operations without synchronization, before recording the end time. 
Then, the measurement overhead is amortized over all the iterations, giving accurate timing estimates. 
When the number of iterations is large enough, the overhead is negligible. 
In our experiments we measure the timing of a large number of computing iterations on a matrix multiplication and use the averaged value of each iteration as the compute time of the matrix multiplication.

\noindent\textbf{Fraction of forward pass spent by matrix multiplication:} 
In Figure~\ref{fig:mm}, we study the fraction of forward pass time spent by matrix multiplication (matmul) operations. 
We do so, by extracting the matmul operations, measuring them, and then comparing with the full forward pass measurement. 
Note that due to the above explained averaging methodology, measurement overhead for matmul operations in this section is negligible.

\input{mm_fig}

First, as seen in \figurename~\ref{fig:tk1cpu}, matmul operations on TK1 CPU take a large portion of forward pass time -- 79.61\%, 96.01\%, 70.15\%, and 91.16\%  for AlexNet, VGGNet, GoogleNet, and ResNet, respectively. 
Note that this also approximates the time taken by CONV and FC layers from Table~\ref{tab:alexnet}, \ref{tab:vgg}, \ref{tab:googlenet}, and \ref{tab:resnet} (81.47\%, 97.98\%, 71.1\%, and 88.57\%).
Second, the trend is similar on TX1 CPU, as depicted in \figurename~\ref{fig:tx1cpu}, except GoogleNet (only about 25\% time spent on matmul operations), which is caused by the particular combination of the architecture of TX1 CPU and GoogleNet as discussed in \S\ref{sec:timing}.
Third, the trend on TK1 and TX1 GPUs is similar to the trend on TX1 CPU, as seen in \figurename~\ref{fig:tk1gpu} and \ref{fig:tx1gpu}. 
One thing to note is that while matmul operations of GoogleNet only take about 20\% of the total time of forward pass, our previous measurement in Table~\ref{tab:googlenet}, showed that CONV and FC layers take about 60\% of the total forward pass time. 
We believe this is because the matmul operations are run without taking into account dependencies, whereas, GoogleNet consists of inception components, each of which has four branches of CONV layers in parallel. 
Before proceeding to next inception component, all four branches of CONV layers have to be competed. 
How to handle such dependencies is part of our future work. 

In summary, for most cases, matmul operations take a large proportion (more than 60\%) of the compute time of a CNN on mobile platforms. 
Thus, we can predict matmul time, to be able to approximately estimate the compute time of a CNN.

\subsection{Modeling}
So far, we have exactly measured matmul time. In this section, we aim to model this time, to be able to predict
the compute time, just from the matrix sizes.
To do so, we benchmark several matrix sizes, as explained below to understand 
the relationship between the size of the matrices and the compute time. 

Given the matmul of $[n\times k]$ and $[k\times m]$ (the number of FLOPs is $n \times m \times k$) performed by a CONV layer, $n$ is the number of kernels, $k$ is the size of a kernel in 3D (width $\times$ height $\times$ depth, where depth is the number of input feature maps), and $m$ is the spatial size (width $\times$ height) of output feature maps.

CNNs follow special rules on these parameters of CONV layers. The number of kernels $n$ is usually a multiple of 16, commonly from 32 to 512. The spatial size of a kernel is commonly $1^2$, $3^2$, $5^2$, $7^2$, or $11^2$. The depth of a kernel is usually the number of kernels in the previous CONV layer and hence also a multiple of 16; except the first CONV layer, where the depth is the number of channels of the input image, typically equal to three. The spatial size of output feature maps of a CNN $m$ gradually reduces; it is common to have $224^2$, $112^2$, $56^2$, $28^2$, $14^2$, or $7^2$, though AlexNet has slightly different ones, \textit{i.e.}, $55^2$, $27^2$ and $13^2$. Based on these typical parameter settings, we carried out experiments on matmul with varying $n$, $m$, and $k$.
The FC layer is currently used in CNNs only as a classifier (\textit{e.g.}, in GoogleNet and ResNet) and thus its compute time is negligible compared to the forward pass. Therefore, we do not consider the size of matrices for FC layers in the modeling.

\textbf{Simple linearity on CPU:} \figurename~\ref{fig:tk1cpu-nmk} and \ref{fig:tx1cpu-nmk} illustrate the performance of matmul on TK1 CPU and TX1 CPU, respectively. The settings of $n$, $m$, and $k$ are: $n=[32,\,64,\,96,\,128,\,256,\,512]$, $m=[7^2,\,14^2,\,28^2,\,56^2,\,112^2]$, and $k=[64\times1^2,\,64\times3^2,\,128\times3^2,\,64\times5^2,\,256\times3^2,\,64\times7^2]$. In each figure, we fix one of three parameters and vary other two; data points are shown as small circles; black circles are labeled with coordinates to highlight the setting of varying parameters. 

From \figurename~\ref{fig:tk1cpu-nm}, \ref{fig:tk1cpu-nk}, and \ref{fig:tk1cpu-mk}, it is observed that the compute time of matmul on TK1 CPU scales linearly with $n$, $m$, and $k$. The linearity can also be observed on TX1 CPU as depicted in \figurename~\ref{fig:tx1cpu-nm}, \ref{fig:tx1cpu-nk}, and \ref{fig:tx1cpu-mk}.
Thus, we have a linear model per CPU device, which predicts the matmul time, given the matrix sizes.

\textbf{Complex linearity on GPU:} \figurename~\ref{fig:tk1gpu-nmk} and \ref{fig:tx1gpu-nmk} illustrate the performance of matmul with varying settings of $n$, $m$, and $k$ on TK1 GPU and TX1 GPU, respectively. The compute time of matmul on GPUs exhibits more complex relationship with $n$, $m$, and $k$.

\figurename~\ref{fig:tk1gpu-nm} mainly depicts the effect of $n$, which is bipartite. 
For all the settings of $m$, the compute time has a monotonic relationship with $n$ from $n=32$ to $128$, except $n=96$ which incurs even longer compute time than $n=128$, while, from $n=128$ to 512, the compute time exhibits a perfect linear relationship with $n$. 
Similar result is also found on TX1 GPU as shown in \figurename~\ref{fig:tx1gpu-nm}. 
Although TX1 GPU has more CUDA cores (256 compare to 192 cores in TK1 GPU) and generally computes matmuls faster than TK1 GPU, it also exhibits this pattern at $n=96$.
This artifact is related to the algorithm that determines how the CUDA cores compute matmul in parallel.
Since cuBLAS is not an open-source library, it is hard to trace the exact reason. 
However, it is indicated \citectl{cublas} that matmul works best if $n$ and $m$ are multiples of 128 on Maxwell architecture (TX1 GPU) and if $n$ is multiple of 256 and $m$ multiple of 192 on Kepler architecture (TK1 GPU). 
This may explain why it behaves differently when $n$ is small. 

For given values of $n$ and $m$, the compute time linearly increases with $k$ on TK1 GPU and TX1 GPU as depicted in \figurename~\ref{fig:tk1gpu-nk} and \ref{fig:tx1gpu-nk}, respectively.
While the compute time increases with $m$ on both TK1 GPU and TX1 GPU as depicted in \figurename~\ref{fig:tk1gpu-mk} and \ref{fig:tx1gpu-mk}, the effect of $m$ is tripartite. The compute time has three separate linear relationships with $k$ (different coefficients), \textit{e.g.}, from $7^2$ to $28^2$, from $28^2$ to $56^2$, and from $56^2$ to $224^2$ on TK1 GPU, as highlighted by different regions in \figurename~\ref{fig:tk1gpu-mk}. In each such region, the compute time on different values of $k$ linearly scales with $m$ at mostly the same coefficient. Moreover, in the middle region (\textit{i.e.}, between $28^2$ and $56^2$ in \figurename~\ref{fig:tk1gpu-mk} and between $14^2$ and $28^2$ in \figurename~\ref{fig:tx1gpu-mk}, the compute time increases with $m$ slower than other two regions. This is especially true on TX1 GPU, where the region is much more flat and tends to plateau. This region should be the transition area, where cuBLAS adopts different schemes based on $m$ and the number of CUDA cores to assign the workload of matmul to CUDA cores. The transition area is different on TK1 GPU and TX1 GPU, mainly because they have different number of CUDA Cores. 

Based on the characteristics discussed above, we are able to model the compute time of matmul on a specific GPU, though we need more data points than that on a CPU.

\subsection{Accuracy}
Based on the measurement, profiling, and modeling of CNNs on mobile devices, we built the modeling tool, \sysname, which estimates the compute time and memory usage for any given CNN. \sysname~ first parses the descriptor of a CNN. Based the type and setting of each layer, it calculates the minimal memory needed to run the CNN.
The memory includes data, parameters, and workspace.
Then, \sysname~ extracts matmuls from the computation of the CNN. Based on the models of TK1 and TX1 on matmul, \emph{i.e.}, the linear fits obtained from \figurename~\ref{fig:tk1cpu-nmk} and \ref{fig:tk1gpu-nmk} for TK1, and \figurename~\ref{fig:tx1cpu-nmk} and \ref{fig:tx1gpu-nmk} for TX1, \sysname~ calculates the compute time of individual matmuls and then uses their summation as the estimate of the compute time of the CNN.

To verify the accuracy of \sysname, we model two CNNs (\textit{i.e.}, NIN \citectl{nin} and VGG19M\footnote{VGG19M is a modified version of VGGNet with more CONV layers. The FC layers in the original VGGNet are replaced by a CONV layer and a POOL layer to reduce memory usage.}) and compare the estimates to the measured memory usage and compute time using Caffe. \figurename~\ref{fig:CNNProfiler-memory} depicts the memory usage of NIN and VGG19M on different processing units. The estimate of memory usage is always less than the actual usage, because the estimate does not take into account the memory usage of Caffe itself, which is framework-dependent. However, it is easy to incorporate that if a specific framework is targeted to perform the CNN computation. Note that the estimate of \sysname~ is accurate on the memory usage of data, parameters, and workspace as discussed in \S\ref{ssec:memory}.

\figurename~\ref{fig:NIN} and \ref{fig:VGG19M} evaluate the accuracy of \sysname's compute time estimation of NIN and VGG19M, respectively. From \figurename~\ref{fig:NIN} and \ref{fig:VGG19M}, we observe that the estimate based on only matmul can approximate the compute time of NIN and VGG19M on both CPUs and GPUs, with more than 78\% accuracy for all the cases. Since matmul generally takes a larger proportion of the compute time on CPUs than on GPUs as discussed in \S \ref{sec:timing}, the estimate on CPUs (up to 94\%) is closer to the actual compute time than on GPUs (up to 84\%). Moreover, more powerful processing unit can perform matmul faster, but the speed up is not the same across all operations.
Therefore, the matmul of a CNN takes a smaller proportion of the compute time on a more powerful processing unit. This explains why the estimate on TK1 CPU (or TK1 GPU) is more accurate than TX1 CPU (TK1 GPU) for the same CNN. 

In summary, \sysname~ can estimate whether and how efficiently a CNN can be run on mobile devices before any deployment. 
It can also help the design of CNNs for resource-constrained mobile devices. When designing a CNN model using \sysname, designers can estimate the resource usage and compute time without implementation and deployment and tune the model to satisfy their specific needs.

\section{Discussion}
\label{sec:discuss}

\sysname~can be extended to support additional mobile platforms by simply profiling matrix multiplication operations on them.
Matrix multiplications of a CNN take most computation (more than 90\% of FLOPs from Table~\ref{tab:alexnet}, \ref{tab:vgg}, \ref{tab:googlenet}, and \ref{tab:resnet}), which commonly takes a dominant proportion of the compute time. 
Thus, matrix multiplication is currently exploited by \sysname~ to estimate the compute time of a CNN. 
To obtain a more precise estimate, additional factors need to be taken into consideration, \textit{e.g.}, memory operations and CNN architectures (stacked or branched). \sysname~ will be enhanced with these features and this will be our future work.

Moreover, we observe that a framework customized for running CNNs on mobile platforms is highly desired. The framework should be optimized for performing the test phase of CNNs and tailored for the characteristics of mobile platforms, \textit{e.g.}, the unified memory architecture.

\section{Conclusion}
\label{conclude}
In this paper, we aim to model the resource requirements of CNNs on mobile devices. 
By deploying several popular CNNs on mobile CPUs and GPUs, we measured and analyzed the performance and resource usage at a layerwise granularity. 
Our findings pointed out the potential ways of optimizing the performance of CNNs on mobile devices. 
As matrix multiplications form the core computations of a CNN, we profiled and modeled matrix multiplications on mobile platforms.
Based on the measurement, profiling, and modeling, we built \sysname~ that can estimate the compute time and memory usage 
of the CNN so as to give insights on whether and how efficiently the CNN can be run on a mobile platform without implementation and deployment. 
Therefore, it is a power tool that helps the design of CNNs for resource-constrained mobile devices.


\bibliographystyle{ACM-Reference-Format}
\bibliography{ref}


\begin{thebibliography}{00}


\ifx \showCODEN    \undefined \def \showCODEN     #1{\unskip}     \fi
\ifx \showDOI      \undefined \def \showDOI       #1{#1}\fi
\ifx \showISBNx    \undefined \def \showISBNx     #1{\unskip}     \fi
\ifx \showISBNxiii \undefined \def \showISBNxiii  #1{\unskip}     \fi
\ifx \showISSN     \undefined \def \showISSN      #1{\unskip}     \fi
\ifx \showLCCN     \undefined \def \showLCCN      #1{\unskip}     \fi
\ifx \shownote     \undefined \def \shownote      #1{#1}          \fi
\ifx \showarticletitle \undefined \def \showarticletitle #1{#1}   \fi
\ifx \showURL      \undefined \def \showURL       {\relax}        \fi
\providecommand\bibfield[2]{#2}
\providecommand\bibinfo[2]{#2}
\providecommand\natexlab[1]{#1}
\providecommand\showeprint[2][]{arXiv:#2}

\bibitem[\protect\citeauthoryear{??}{caf}{}]%
        {caffe:long}
\bibinfo{booktitle}{{\em Caffe}}.
\newblock
\newblock
\shownote{\url{http://caffe.berkeleyvision.org/}.}


\bibitem[\protect\citeauthoryear{??}{cub}{}]%
        {cublas:long}
\bibinfo{booktitle}{{\em cuBLAS}}.
\newblock
\newblock
\shownote{\url{https://developer.nvidia.com/cublas}.}


\bibitem[\protect\citeauthoryear{??}{cud}{a}]%
        {cuda-guide:long}
\bibinfo{booktitle}{{\em CUDA C Programming Guide}}.
\newblock
\newblock
\shownote{\url{https://docs.nvidia.com/cuda/}.}


\bibitem[\protect\citeauthoryear{??}{cud}{b}]%
        {cudnn:long}
\bibinfo{booktitle}{{\em cuDNN}}.
\newblock
\newblock
\shownote{\url{https://developer.nvidia.com/cudnn/}.}


\bibitem[\protect\citeauthoryear{??}{ope}{}]%
        {openblas:long}
\bibinfo{booktitle}{{\em OpenBLAS}}.
\newblock
\newblock
\shownote{\url{http://www.openblas.net/}.}


\bibitem[\protect\citeauthoryear{??}{ten}{}]%
        {tensorflow:long}
\bibinfo{booktitle}{{\em TensorFlow}}.
\newblock
\newblock
\shownote{\url{http://www.tensorflow.org/}.}


\bibitem[\protect\citeauthoryear{??}{the}{}]%
        {theano:long}
\bibinfo{booktitle}{{\em Theano}}.
\newblock
\newblock
\shownote{\url{http://deeplearning.net/software/theano/}.}


\bibitem[\protect\citeauthoryear{??}{tor}{}]%
        {torch:long}
\bibinfo{booktitle}{{\em Torch}}.
\newblock
\newblock
\shownote{\url{http://torch.ch/}.}


\bibitem[\protect\citeauthoryear{Canziani, Paszke, and Culurciello}{Canziani
  et~al\mbox{.}}{2016}]%
        {canziani2016analysis:long}
\bibfield{author}{\bibinfo{person}{Alfredo Canziani}, \bibinfo{person}{Adam
  Paszke}, {and} \bibinfo{person}{Eugenio Culurciello}.}
  \bibinfo{year}{2016}\natexlab{}.
\newblock \showarticletitle{An Analysis of Deep Neural Network Models for
  Practical Applications}.
\newblock \bibinfo{journal}{{\em arXiv preprint arXiv:1605.07678\/}}
  (\bibinfo{year}{2016}).
\newblock


\bibitem[\protect\citeauthoryear{Gong, Liu, Yang, and Bourdev}{Gong
  et~al\mbox{.}}{2014}]%
        {gong2014compressing:long}
\bibfield{author}{\bibinfo{person}{Yunchao Gong}, \bibinfo{person}{Liu Liu},
  \bibinfo{person}{Ming Yang}, {and} \bibinfo{person}{Lubomir Bourdev}.}
  \bibinfo{year}{2014}\natexlab{}.
\newblock \showarticletitle{Compressing deep convolutional networks using
  vector quantization}.
\newblock \bibinfo{journal}{{\em arXiv preprint arXiv:1412.6115\/}}
  (\bibinfo{year}{2014}).
\newblock


\bibitem[\protect\citeauthoryear{Han, Shen, Philipose, Agarwal, Wolman, and
  Krishnamurthy}{Han et~al\mbox{.}}{2016}]%
        {han2016mcdnn:long}
\bibfield{author}{\bibinfo{person}{Seungyeop Han}, \bibinfo{person}{Haichen
  Shen}, \bibinfo{person}{Matthai Philipose}, \bibinfo{person}{Sharad Agarwal},
  \bibinfo{person}{Alec Wolman}, {and} \bibinfo{person}{Arvind Krishnamurthy}.}
  \bibinfo{year}{2016}\natexlab{}.
\newblock \showarticletitle{MCDNN: An Approximation-Based Execution Framework
  for Deep Stream Processing Under Resource Constraints}. In
  \bibinfo{booktitle}{{\em International Conference on Mobile Systems,
  Applications, and Services}} {\em (\bibinfo{series}{MobiSys'16})}.
\newblock


\bibitem[\protect\citeauthoryear{He, Zhang, Ren, and Sun}{He
  et~al\mbox{.}}{2016}]%
        {resnet:long}
\bibfield{author}{\bibinfo{person}{Kaiming He}, \bibinfo{person}{Xiangyu
  Zhang}, \bibinfo{person}{Shaoqing Ren}, {and} \bibinfo{person}{Jian Sun}.}
  \bibinfo{year}{2016}\natexlab{}.
\newblock \showarticletitle{Deep Residual Learning for Image Recognition}. In
  \bibinfo{booktitle}{{\em IEEE Conference on Computer Vision and Pattern
  Recognition}} {\em (\bibinfo{series}{CVPR'16})}.
\newblock


\bibitem[\protect\citeauthoryear{Iandola, Han, Moskewicz, Ashraf, Dally, and
  Keutzer}{Iandola et~al\mbox{.}}{2016}]%
        {iandola2016squeezenet:long}
\bibfield{author}{\bibinfo{person}{Forrest~N Iandola}, \bibinfo{person}{Song
  Han}, \bibinfo{person}{Matthew~W Moskewicz}, \bibinfo{person}{Khalid Ashraf},
  \bibinfo{person}{William~J Dally}, {and} \bibinfo{person}{Kurt Keutzer}.}
  \bibinfo{year}{2016}\natexlab{}.
\newblock \showarticletitle{SqueezeNet: AlexNet-level accuracy with 50x fewer
  parameters and {$<$}0.5{MB} model size}.
\newblock \bibinfo{journal}{{\em arXiv preprint arXiv:1602.07360\/}}
  (\bibinfo{year}{2016}).
\newblock


\bibitem[\protect\citeauthoryear{Ioffe and Szegedy}{Ioffe and Szegedy}{2015}]%
        {batchnorm:long}
\bibfield{author}{\bibinfo{person}{Sergey Ioffe} {and}
  \bibinfo{person}{Christian Szegedy}.} \bibinfo{year}{2015}\natexlab{}.
\newblock \showarticletitle{Batch Normalization: Accelerating Deep Network
  Training by Reducing Internal Covariate Shift}. In \bibinfo{booktitle}{{\em
  International Conference on Machine Learning}} {\em
  (\bibinfo{series}{ICML'15})}.
\newblock


\bibitem[\protect\citeauthoryear{Jia, Shelhamer, Donahue, Karayev, Long,
  Girshick, Guadarrama, and Darrell}{Jia et~al\mbox{.}}{2014}]%
        {jia2014caffe:long}
\bibfield{author}{\bibinfo{person}{Yangqing Jia}, \bibinfo{person}{Evan
  Shelhamer}, \bibinfo{person}{Jeff Donahue}, \bibinfo{person}{Sergey Karayev},
  \bibinfo{person}{Jonathan Long}, \bibinfo{person}{Ross Girshick},
  \bibinfo{person}{Sergio Guadarrama}, {and} \bibinfo{person}{Trevor Darrell}.}
  \bibinfo{year}{2014}\natexlab{}.
\newblock \showarticletitle{Caffe: Convolutional Architecture for Fast Feature
  Embedding}. In \bibinfo{booktitle}{{\em ACM International Conference on
  Multimedia}} {\em (\bibinfo{series}{MM'14})}.
\newblock


\bibitem[\protect\citeauthoryear{Kim, Park, Yoo, Choi, Yang, and Shin}{Kim
  et~al\mbox{.}}{2016}]%
        {kim2016compression:long}
\bibfield{author}{\bibinfo{person}{Yong-Deok Kim}, \bibinfo{person}{Eunhyeok
  Park}, \bibinfo{person}{Sungjoo Yoo}, \bibinfo{person}{Taelim Choi},
  \bibinfo{person}{Lu Yang}, {and} \bibinfo{person}{Dongjun Shin}.}
  \bibinfo{year}{2016}\natexlab{}.
\newblock \showarticletitle{Compression of Deep Convolutional Neural Networks
  for Fast and Low Power Mobile Applications}. In \bibinfo{booktitle}{{\em
  International Conference on Learning Representations}} {\em
  (\bibinfo{series}{ICLR'16})}.
\newblock


\bibitem[\protect\citeauthoryear{Krizhevsky, Sutskever, and Hinton}{Krizhevsky
  et~al\mbox{.}}{2012}]%
        {alexnet:long}
\bibfield{author}{\bibinfo{person}{Alex Krizhevsky}, \bibinfo{person}{Ilya
  Sutskever}, {and} \bibinfo{person}{Geoffrey~E Hinton}.}
  \bibinfo{year}{2012}\natexlab{}.
\newblock \showarticletitle{Imagenet Classification with Deep Convolutional
  Neural Networks}. In \bibinfo{booktitle}{{\em Neural Information Processing
  Systems Conference}} {\em (\bibinfo{series}{NIPS'12})}.
\newblock


\bibitem[\protect\citeauthoryear{Lane, Bhattacharya, Georgiev, Forlivesi, Jiao,
  Qendro, and Kawsar}{Lane et~al\mbox{.}}{2016}]%
        {lane2016deepx:long}
\bibfield{author}{\bibinfo{person}{Nicholas~D Lane}, \bibinfo{person}{Sourav
  Bhattacharya}, \bibinfo{person}{Petko Georgiev}, \bibinfo{person}{Claudio
  Forlivesi}, \bibinfo{person}{Lei Jiao}, \bibinfo{person}{Lorena Qendro},
  {and} \bibinfo{person}{Fahim Kawsar}.} \bibinfo{year}{2016}\natexlab{}.
\newblock \showarticletitle{Deepx: A Software Accelerator for Low-Power Deep
  Learning Inference on Mobile Devices}. In \bibinfo{booktitle}{{\em
  International Conference on Information Processing in Sensor Networks}} {\em
  (\bibinfo{series}{IPSN'16})}.
\newblock


\bibitem[\protect\citeauthoryear{Lane, Bhattacharya, Georgiev, Forlivesi, and
  Kawsar}{Lane et~al\mbox{.}}{2015}]%
        {lane2015early:long}
\bibfield{author}{\bibinfo{person}{Nicholas~D Lane}, \bibinfo{person}{Sourav
  Bhattacharya}, \bibinfo{person}{Petko Georgiev}, \bibinfo{person}{Claudio
  Forlivesi}, {and} \bibinfo{person}{Fahim Kawsar}.}
  \bibinfo{year}{2015}\natexlab{}.
\newblock \showarticletitle{An Early Resource Characterization of Deep Learning
  on Wearables, Smartphones and Internet-of-Things Devices}. In
  \bibinfo{booktitle}{{\em International Workshop on Internet of Things towards
  Applications}} {\em (\bibinfo{series}{IoT-App'15})}.
\newblock


\bibitem[\protect\citeauthoryear{Lin, Chen, and Yan}{Lin et~al\mbox{.}}{2014}]%
        {nin:long}
\bibfield{author}{\bibinfo{person}{Min Lin}, \bibinfo{person}{Qiang Chen},
  {and} \bibinfo{person}{Shuicheng Yan}.} \bibinfo{year}{2014}\natexlab{}.
\newblock \showarticletitle{Network in Network}. In \bibinfo{booktitle}{{\em
  International Conference on Learning Representations}} {\em
  (\bibinfo{series}{ICLR'14})}.
\newblock


\bibitem[\protect\citeauthoryear{Simonyan and Zisserman}{Simonyan and
  Zisserman}{2015}]%
        {vgg:long}
\bibfield{author}{\bibinfo{person}{Karen Simonyan} {and}
  \bibinfo{person}{Andrew Zisserman}.} \bibinfo{year}{2015}\natexlab{}.
\newblock \showarticletitle{Very Deep Convolutional Networks for Large-Scale
  Image recognition}. In \bibinfo{booktitle}{{\em International Conference on
  Learning Representations}} {\em (\bibinfo{series}{ICLR'15})}.
\newblock


\bibitem[\protect\citeauthoryear{Szegedy, Liu, Jia, Sermanet, Reed, Anguelov,
  Erhan, Vanhoucke, and Rabinovich}{Szegedy et~al\mbox{.}}{2015}]%
        {google:long}
\bibfield{author}{\bibinfo{person}{Christian Szegedy}, \bibinfo{person}{Wei
  Liu}, \bibinfo{person}{Yangqing Jia}, \bibinfo{person}{Pierre Sermanet},
  \bibinfo{person}{Scott Reed}, \bibinfo{person}{Dragomir Anguelov},
  \bibinfo{person}{Dumitru Erhan}, \bibinfo{person}{Vincent Vanhoucke}, {and}
  \bibinfo{person}{Andrew Rabinovich}.} \bibinfo{year}{2015}\natexlab{}.
\newblock \showarticletitle{Going Deeper with Convolutions}. In
  \bibinfo{booktitle}{{\em IEEE Conference on Computer Vision and Pattern
  Recognition}} {\em (\bibinfo{series}{CVPR'15})}.
\newblock


\bibitem[\protect\citeauthoryear{Wu, Leng, Wang, Hu, and Cheng}{Wu
  et~al\mbox{.}}{2016}]%
        {wu2016quantized:long}
\bibfield{author}{\bibinfo{person}{Jiaxiang Wu}, \bibinfo{person}{Cong Leng},
  \bibinfo{person}{Yuhang Wang}, \bibinfo{person}{Qinghao Hu}, {and}
  \bibinfo{person}{Jian Cheng}.} \bibinfo{year}{2016}\natexlab{}.
\newblock \showarticletitle{Quantized Convolutional Neural Networks for Mobile
  Devices}. In \bibinfo{booktitle}{{\em IEEE Conference on Computer Vision and
  Pattern Recognition}} {\em (\bibinfo{series}{CVPR'16})}.
\newblock


\end{thebibliography}



\end{document}